\documentclass[letterpaper, 10 pt, conference]{ieeeconf}  

\IEEEoverridecommandlockouts                              

\usepackage{latexsym}
\usepackage{amsmath}
\usepackage{amssymb}
\usepackage{epsfig}
\usepackage{bbm}
\usepackage{caption}
\usepackage{dblfloatfix}
\usepackage{subcaption}
\usepackage[table]{xcolor}

\usepackage{cite}

\usepackage{algorithm,algpseudocode}

\usepackage[export]{adjustbox}
\usepackage{verbatim}
\usepackage{hyperref}
\hypersetup{%
  breaklinks,
  bookmarksnumbered=true,
  bookmarksopen=true,
  colorlinks=true,
  linkcolor=black,
  urlcolor=black,
  citecolor=black
}
\usepackage{xcolor}
\usepackage{subfiles}

\begin{document}
\bstctlcite{IEEEexample:BSTcontrol}
%
\title{The POLAR Traverse Dataset: A Dataset of Stereo Camera Images Simulating Traverses across Lunar Polar Terrain under Extreme Lighting Conditions}


\author{Margaret Hansen$^{1*}$
\thanks{$^{1}$ Robotics Institute, Carnegie Mellon University, Pittsburgh, PA, USA.\newline$^{2}$ NASA Ames Research Center, Mountain View, CA, USA.\newline $^{*}$ Corresponding author. margareh@cs.cmu.edu.}%
, Uland Wong$^{2}$, Terrence Fong$^{2}$
}

\maketitle
\IEEEpeerreviewmaketitle



\begin{abstract}
We present the POLAR Traverse Dataset: a dataset of high-fidelity stereo pair images of lunar-like terrain under polar lighting conditions designed to simulate a straight-line traverse. Images from individual traverses with different camera heights and pitches were recorded at 1 m intervals by moving a suspended stereo bar across a test bed filled with regolith simulant and shaped to mimic lunar south polar terrain. Ground truth geometry and camera position information was also recorded. This dataset is intended for developing and testing software algorithms that rely on stereo or monocular camera images, such as visual odometry, for use in the lunar polar environment, as well as to provide insight into the expected lighting conditions in lunar polar regions.
\end{abstract}

\section{Introduction}

The lunar south polar region is of particular interest to upcoming NASA missions such as the Volatiles Investigating Polar Exploration Rover (VIPER) due to the existence of cold traps, potential sources for stable water ice \cite{ennico-smith_viper_2022}. This region is characterized by extreme lighting conditions that can be troublesome for both human and robotic explorers, key among them the combination of long shadows adjacent to areas of bright, direct sunlight. As a result, future exploration in these regions faces the challenge of navigating through such a visually confusing environment. Few previous missions have targeted this region for landing, meaning relatively little data is available for developing and testing perception algorithms for robotic exploration and navigation tasks. Inspired by the POLAR Stereo Dataset \cite{wong_characterization_2016}, which provides static stereo images of lunar-like terrain relevant to these lighting conditions, the POLAR Traverse Dataset is intended as a means of filling this gap with the goal of providing high-fidelity images from simulated straight-line traverses across a lunar-like environment under polar lighting conditions. The full dataset and associated documentation can be downloaded at \href{https://ti.arc.nasa.gov/dataset/PolarTrav/}{\color{blue}https://ti.arc.nasa.gov/dataset/PolarTrav/}.

\section{Background}


\begin{figure}
    \centering
    \includegraphics[width=0.45\textwidth]{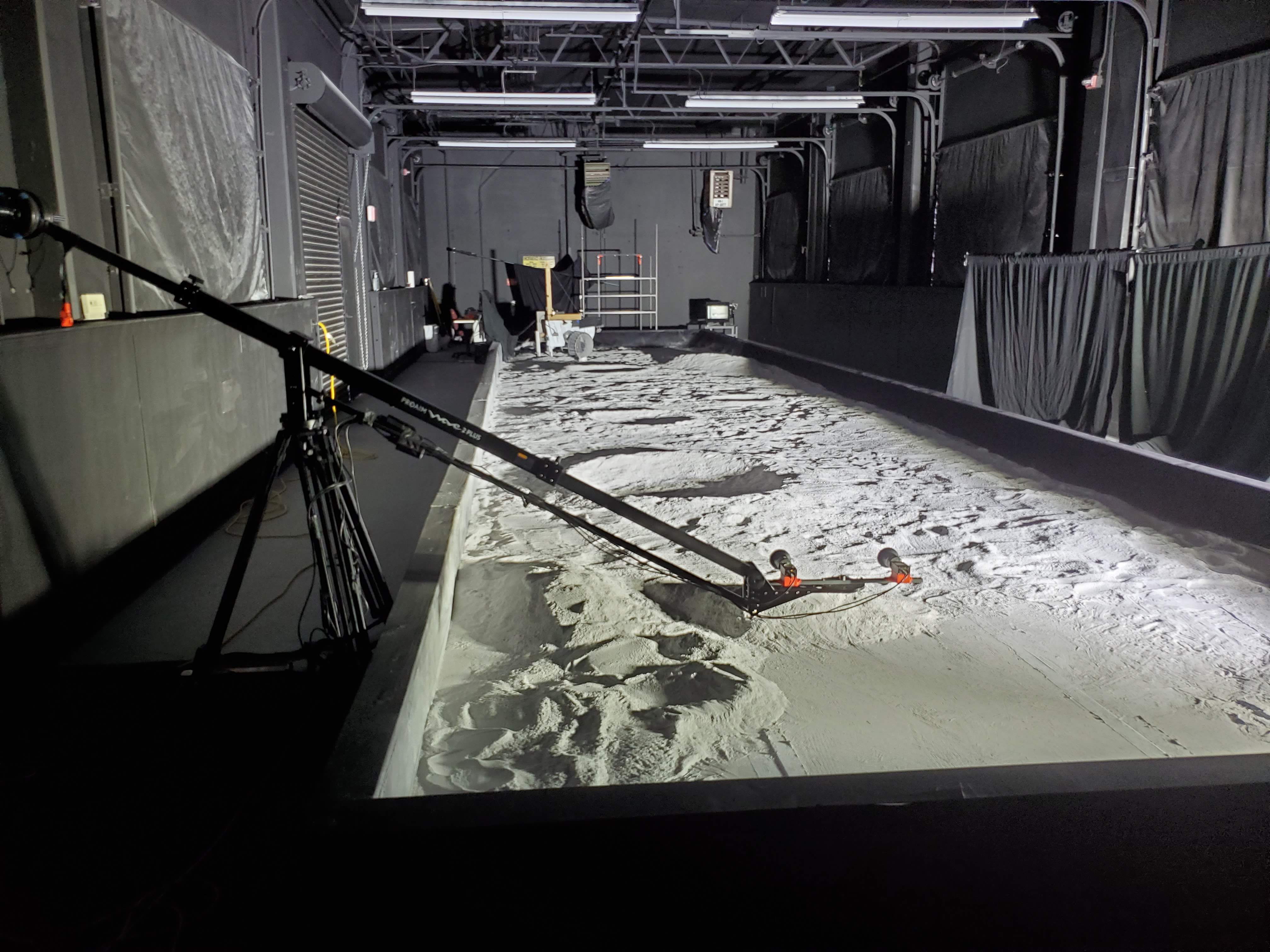}
    \caption{Hardware setup extended over SSERVI test bed with lunar terrain and lighting.}
    \label{fig:camera_rig}
\end{figure}

\subsection{Future Lunar Exploration}

A stated goal of NASA's Moon to Mars program is to characterise water deposits on the moon as a resource to be extracted and used \emph{in situ} \cite{moon2mars}. The upcoming VIPER rover mission, targeted fro launch in late 2024 \cite{viper_rover}, is tasked with better characterizing the distribution of water ice on the moon and will build resource maps towards this goal \cite{shirley_viper_2022}. Likewise, the science objectives for the Artemis III mission include characterizing lunar volatiles \cite{ArtemisScienceReport2021}.

Water ice is stable in the subsurface and on the surface in permanently shadowed regions (PSRs) near the lunar poles \cite{fisher_evidence_2017}, with micro cold traps being of particular interest as they are smaller, easier to access, and more abundant than the larger PSRs \cite{hayne_micro_2020}. These areas exist in the polar regions due to the low incidence angle of incoming sunlight, which poses challenges for robotic and human explorers while at the same time making these regions a prime target for exploration because of the potential existence of water ice on the surface in permanently shadowed regions \cite{ennico-smith_viper_2022}. Contending with these extremes is necessary for NASA's planned lunar exploration activities, such as robotic navigation through polar regions \cite{ennico-smith_viper_2022} and astronaut EVAs \cite{crues_approaches_2023, null_identification_2023}.



\subsection{Lunar Polar Lighting Conditions}

The extreme lighting conditions found at the lunar poles simultaneously make simulation difficult in a terrestrial test bed and result in degraded performance of standard image-based perception algorithms. The low incidence of incoming sunlight ($\sim$1.5\textdegree\, elevation at the south pole) results in high visual contrast between brightly lit areas and long, large shadows \cite{crues_approaches_2023}. In addition, the lack of atmosphere on the moon prevents any diffusion of light into shadowed areas from occurring as it would on Earth \cite{null_identification_2023}. Typical approaches to stereo image processing, including multi-view stereo, used to generate terrain models of planetary surfaces rely on feature extraction and matching and are known to fail in regions with high amounts of shadow \cite{stefano_fast_2004}. Development of more robust perception algorithms is necessary for automated exploration of regions with such extreme lighting. The POLAR Traverse Dataset is intended to provide example imagery taken under such conditions with high optical fidelity for this type of development work.

\begin{figure}[!tbp]
    \centering
    \includegraphics[width=0.45\textwidth]{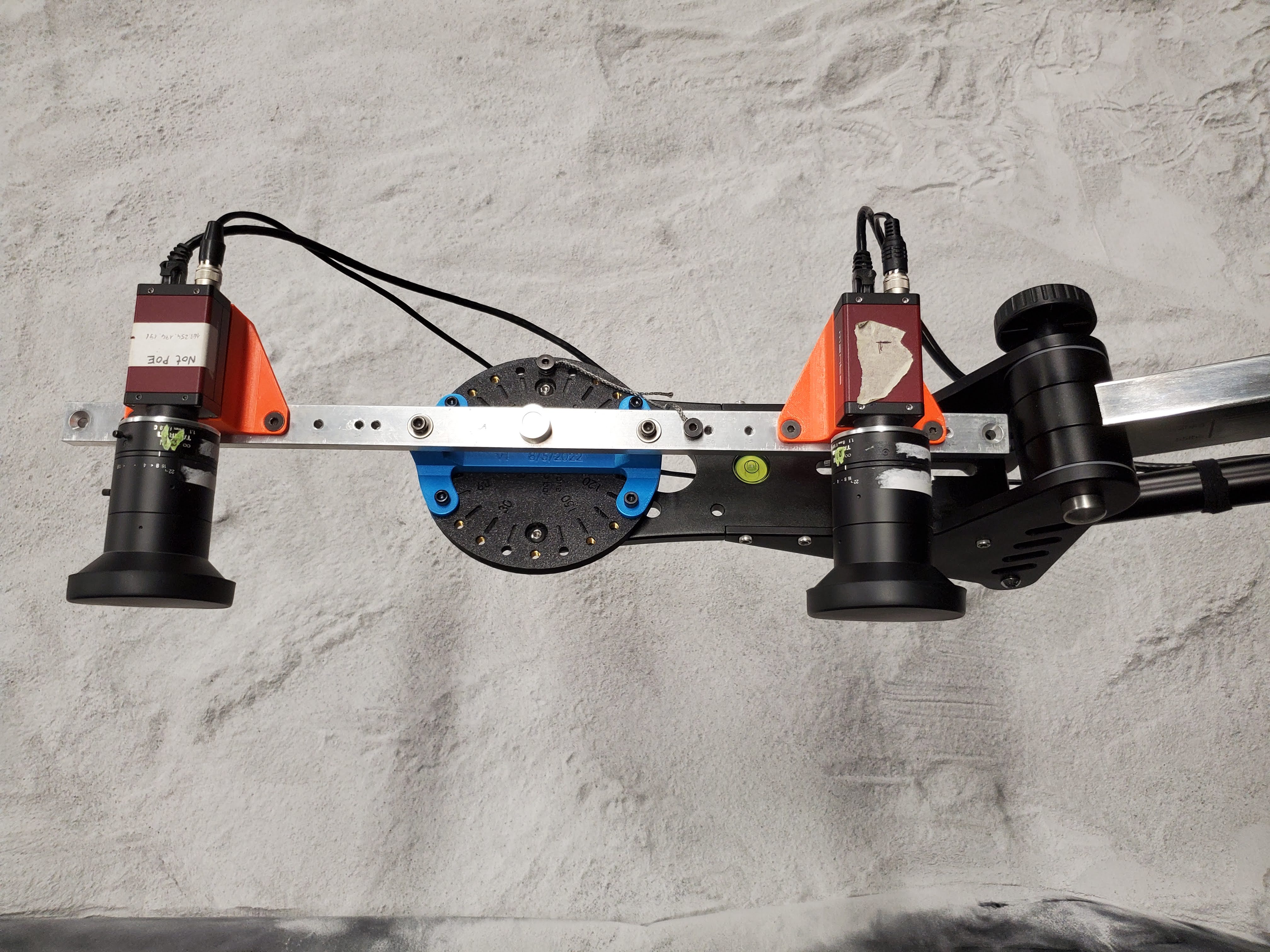}
    \caption{Closeup of stereo bar attached to end of jib arm with pitch mount (blue).}
    \label{fig:hardware_setup}
\end{figure}

\section{Setup}

\subsection{Scene Construction}

The test bed used for data collection, shown in Figure~\ref{fig:camera_rig}, measures 18.6 m long by 4 m wide and has been provided by the Solar System Exploration Research Virtual Institute (SSERVI) \cite{lunar_lab}. Inside the test bed was a lunar-like scene with a portion of a human-made structure stationed on top of the terrain at one end, which can be seen in the ground truth heightmap in Figure~\ref{fig:ground_truth}. The terrain was constructed to reflect planetary science knowledge about the size, distribution, and depth of craters in the south polar region of the moon. It contains four craters of varying diameter, along with uneven terrain and a few larger mounds. The scene was lit by lights set up near the corners of the test bed and at a low angle to the terrain ($\sim$2-3\textdegree) to simulate the low elevations at which sunlight is incident on lunar polar terrain.

The simulant used to construct the scene inside the test bed is a modified version of the LHS-1 simulant produced by Exolith Labs \cite{isachenkov_char_2022}. LHS-1 is designed to mimic the mineral composition and particle size distribution of lunar highlands regolith \cite{isachenkov_char_2022}, the type found in the south polar regions \cite{long-fox_applicability_2022}. The modified version excludes the trace minerals used in LHS-1 for cost reasons due to the need to produce it in bulk for a test bed of this size. The composition of mLHS-1 differs from LHS-1 by around 1\%, while its optical properties differ by much less than 1\%.

\subsection{Hardware}

Two 16 mm diagonal Allied Vision Manta G-419 \cite{manta} cameras were each equipped with a Tamron M111FM08 8mm f/1.8 lens \cite{tamron} set at f8 and mounted to a stereo bar with a 40 cm baseline. The stereo bar was attached to the end of a camera jib arm for suspension over the test bed, as shown in Figure~\ref{fig:camera_rig}. Interchangeable pitch mounts were used when attaching the stereo bar to the jib arm to simulate three camera pitch angles (14°, 25°, 35°), each of which was used for certain data collection iterations. A closeup of the stereo bar setup is shown in Figure~\ref{fig:hardware_setup}.

Geometric calibration was performed with a 9 x 12 checkerboard with 60 mm squares prior to data collection. 71 calibration image pairs were collected with an exposure time of 100 ms and with the overhead lights turned on. The calibration was processed using MATLAB's Stereo Camera Calibrator app \cite{matlab}, during which 8 image pairs were removed due to missed detection of checkerboard corners and 1 was manually removed as an outlier. The final calibration produced an average reprojection error of 0.111 pixels and is provided as part of the dataset.

\section{Collection Procedure}

\subsection{Traverse Images}

\begin{figure*}[!htbp]
    \centering
    \includegraphics[width=0.9\textwidth]{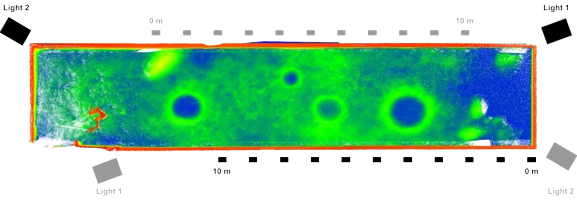}
    \caption{Top-down view of the scene inside the test bed from the processed ground truth scan, displaying lunar-like terrain and the human-made object (lower left corner). Color denotes height (blue is lowest, red is highest). Approximate light locations and camera positions are shown for each viewing direction (black is forward, grey is reverse).}
    \label{fig:ground_truth}
\end{figure*}

To simulate traverses across the lunar surface, the cameras were suspended above the terrain facing along the length of the test bed. Starting at one end, stereo pairs of full resolution (2048 x 2048) images were recorded every 1 m along the test bed by moving the camera jib base to the next position, stopping, collecting and verifying images with multiple exposure times, and proceeding. Images were recorded using the following exposure times at each position (all in ms): 1, 5, 10, 25, 50, 75, 100, 150, 200, 250, 300, 350, 400, 450, 500. Collecting images with multiple exposure times was done to provide the ability to test how exposure impacts the performance of perception algorithms such as stereo or visual odometry. This procedure was repeated for 10 m along the test bed, resulting in 11 camera positions (0 m – 10 m) for one traverse. Approximate positions for the camera jib arm along the length of the test bed are shown in Figure~\ref{fig:ground_truth}.

\begin{table}[!htbp]
    \centering
    \begin{tabular}[t]{c|c|c}
        View & Viewing Direction & Lighting Angle \\
         \hline
        1 & Forward & 20\textdegree \\
        2 & Forward & 150\textdegree \\
        3 & Reverse & 20\textdegree \\
        4 & Reverse & 150\textdegree \\
         \hline
    \end{tabular}
    \caption{Parameters for each terrain view.}\label{tab:environs}
\end{table}

The collection process was repeated for a total of 24 traverses with varying camera extrinsic parameters and scene conditions. The variable scene conditions (termed ``terrain views'') are the direction in which the cameras were facing along the test bed (the ``viewing direction'') and the angle of incidence of simulated sunlight relative to this direction (20°, 150°). The four terrain views included in this dataset are shown in Table~\ref{tab:environs}. Viewing the same scene from two directions is used to provide some variation in the observed terrain in lieu of the time-consuming process of building an entirely new scene.

The variable camera extrinsics (denoting ``traverses'' within a terrain view) were the camera height, set to either 1.3 m or 0.65 m, in addition to the three camera pitch angles; these combinations are listed in Table~\ref{tab:traverses}. Data was collected for all 6 traverses for each of the 4 terrain views.

\begin{table}[!htbp]
    \centering
    \begin{tabular}[t]{c|c|c}
         Traverse & Camera Height & Camera Pitch \\
         \hline
        1 & 1.3 m & 14\textdegree \\
        2 & 1.3 m & 25\textdegree \\
        3 & 1.3 m & 35\textdegree \\
        4 & 0.65 m & 14\textdegree \\
        5 & 0.65 m & 25\textdegree \\
        6 & 0.65 m & 35\textdegree \\
         \hline
    \end{tabular}
    \caption{Parameters for each traverse.}\label{tab:traverses}
\end{table}

\subsection{Ground Truth}

Four high resolution (8 mm) LiDAR scans of the test bed terrain were collected to provide ground truth geometry information, with two scans taken on each side of the test bed at different distances along the length of the bed to provide roughly evenly distributed information across the majority of the scene. These four scans are included in the dataset in raw form after adjusting each scan’s height values to have a minimum value of zero. In addition, a composite scan was generated by registering the four scans to one another and applying some minimal processing; this scan is also included in the dataset as the full ground truth point cloud and is shown in Figure~\ref{fig:ground_truth}. The processed scan is cropped to the test bed, while the raw scans contain points from the external environment.

\subsection{Pose Estimation}

Camera positions for both cameras were initially approximated based on knowledge of the camera height, pitch, and length along the test bed at which images were recorded, in addition to the stereo baseline. These approximate positions are included in the dataset, along with refined positions generated using the COLMAP structure-from-motion pipeline \cite{schonberger_structure--motion_2016}. The images were downsampled to half-resolution (1024 x 1024) to speed up COLMAP's processing time, and the default parameters were adjusted through trial and error to achieve convergence. In addition, approximate poses were provided to COLMAP as initial estimations. Since COLMAP relies on feature extraction and matching, analyzing its performance at scene reconstruction can provide some insight into the difficulties these types of algorithms face in lunar polar lighting conditions. The reconstruction error associated with the COLMAP reconstruction for each terrain view is provided in Table~\ref{tab:pose_error}, while Figure~\ref{fig:colmap_dense} provides an example dense reconstruction from COLMAP.


\begin{figure*}[!htbp]
    \centering
    \begin{subfigure}{0.45\textwidth}
        \includegraphics[width=\textwidth]{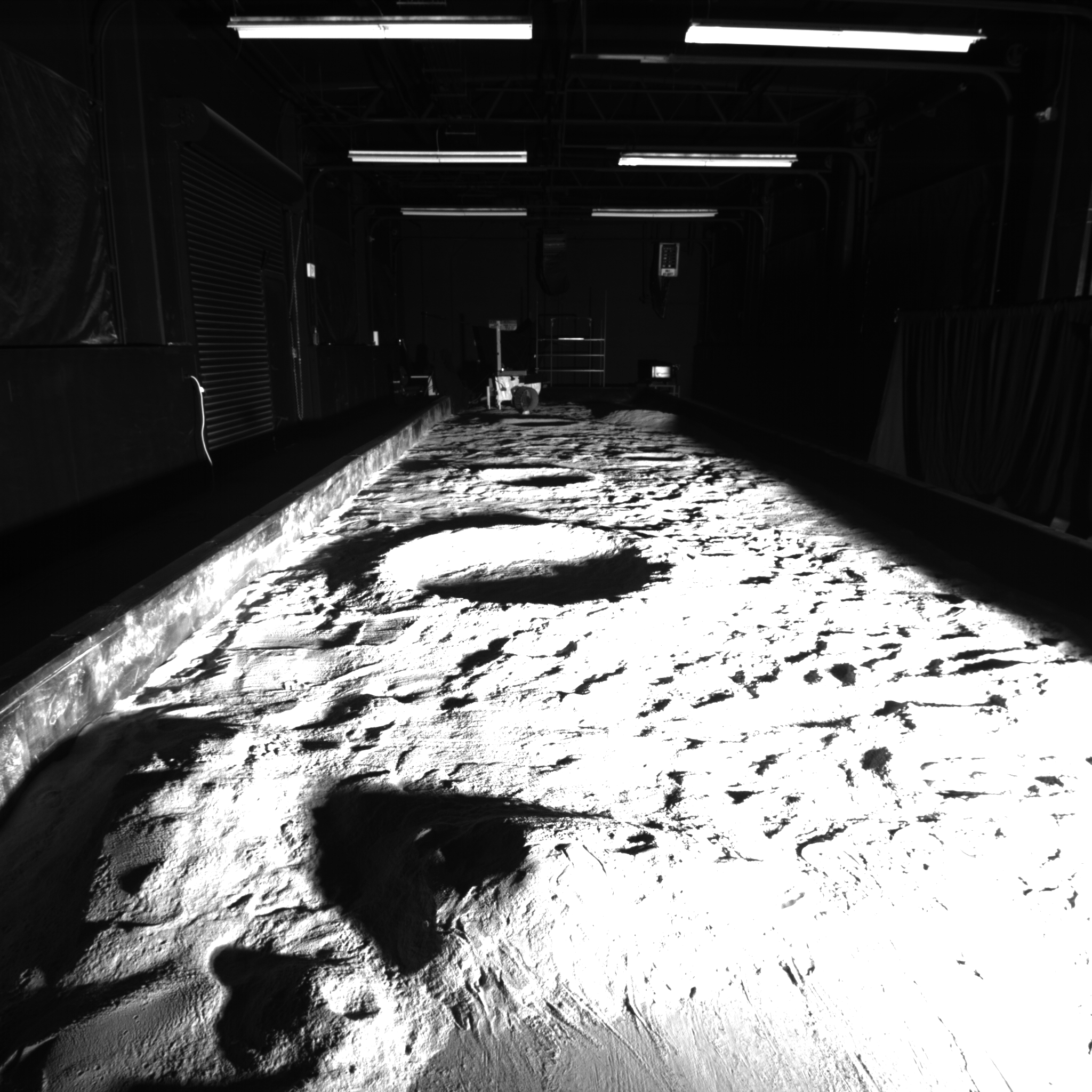}
    \end{subfigure}
    \hfill
    \begin{subfigure}{0.45\textwidth}
        \includegraphics[width=\textwidth]{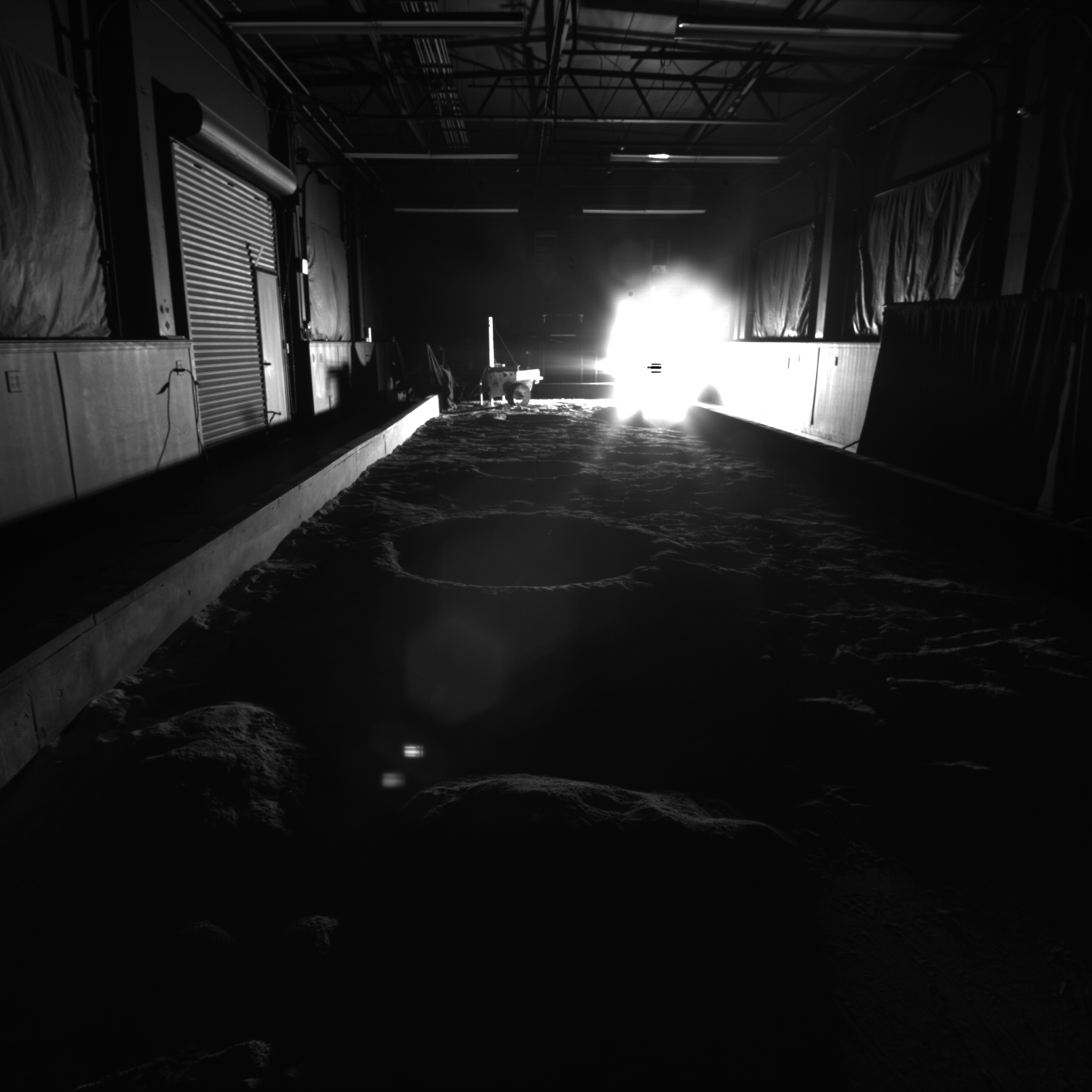}
    \end{subfigure}
    \\
    \vspace{2mm}
    \begin{subfigure}{0.45\textwidth}
        \includegraphics[width=\textwidth]{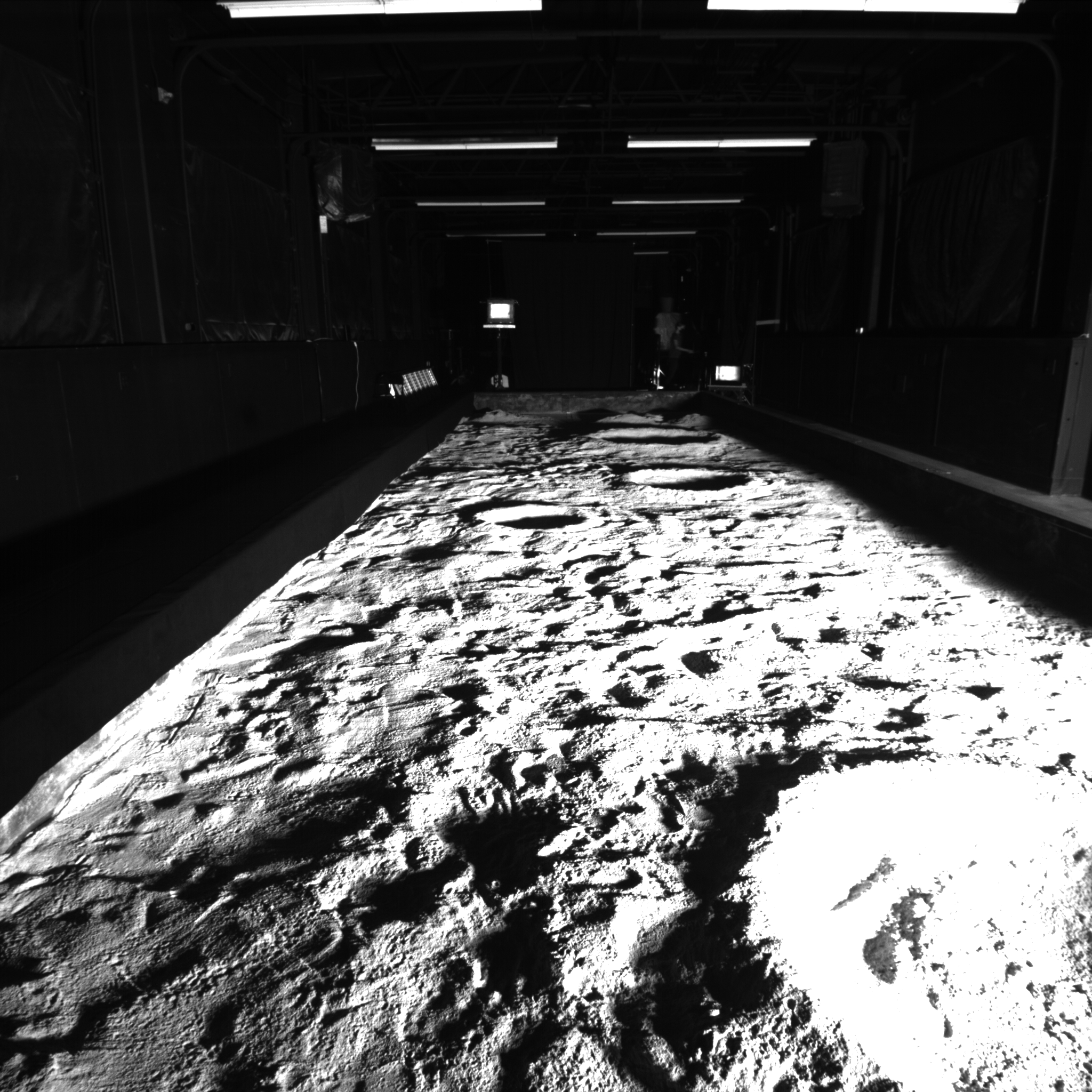}
    \end{subfigure}
    \hfill
    \begin{subfigure}{0.45\textwidth}
        \includegraphics[width=\textwidth]{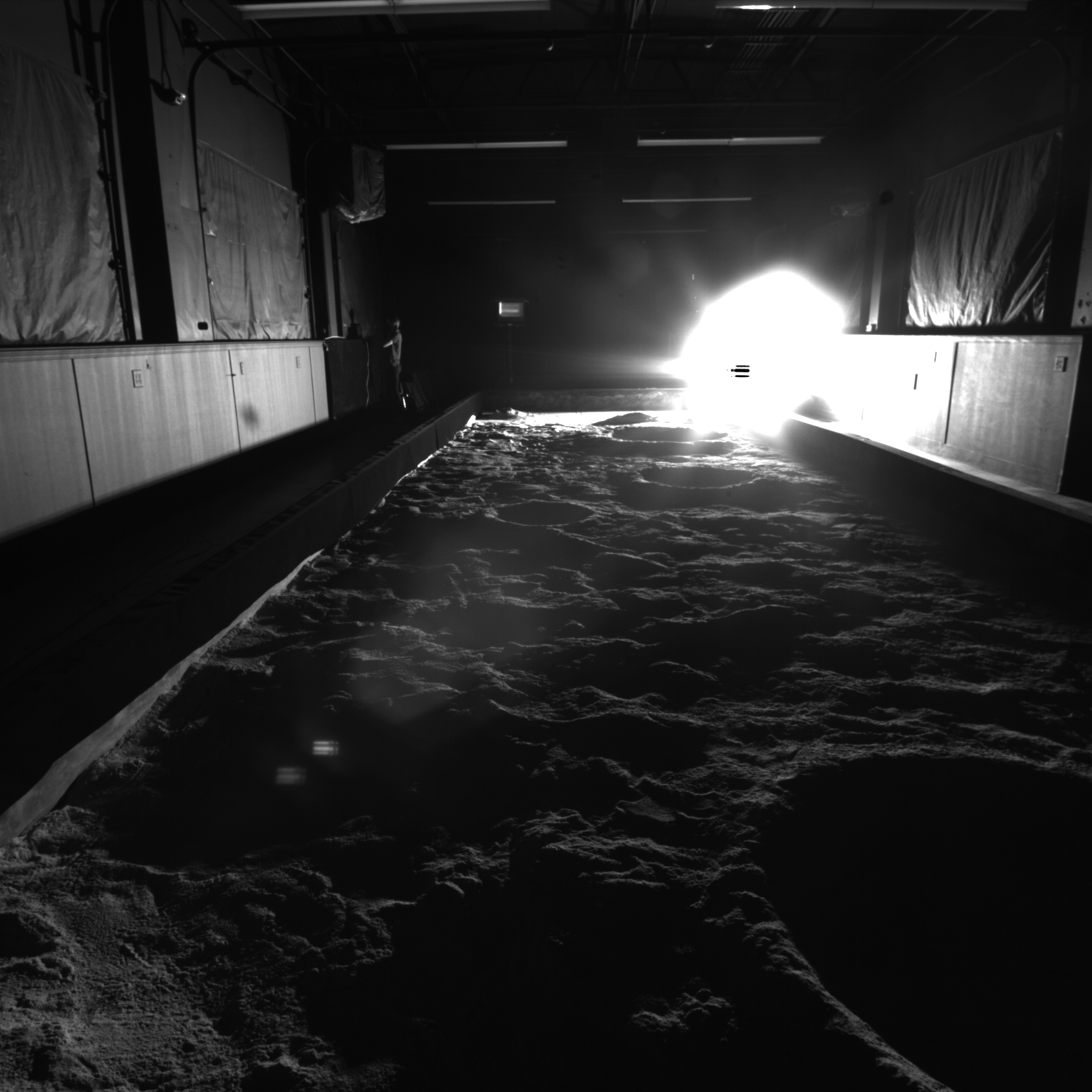}
    \end{subfigure}
    \caption{Example left camera images from the first sampling location for the two forward-facing terrain views (1, 2) available in the POLAR Traverse Dataset. Both images were taken with the same extrinsic parameters (14\textdegree\, camera pitch, 1.3 m above the terrain, 50 ms exposure time).}
    \label{fig:example_images_fwd}
\end{figure*}

\section{Results \& Discussion}

A total of 3,960 stereo pairs of images were recorded across 4 different terrain views with 6 traverses per view. Each traverse includes 11 camera positions along the length of the test bed, with images taken using 15 exposure times at each position. The dataset can be downloaded as a whole (13.4 GB compressed) or as individual terrain view datasets ($\sim$3.3 GB compressed each), with the 4 latter downloads including all 6 traverses for a specific view. The individual datasets do not include the ground truth scan, which is also available as a separate download. Example images from each of the forward and reverse terrain views can be seen in Figure~\ref{fig:example_images_fwd}.


\subsection{Multi-View Stereo Reconstruction}\label{ssec:colmap}

As a multi-view stereo algorithm dependent on feature extraction and matching, COLMAP was expected to perform poorly under lunar polar lighting conditions. Analyzing its performance when producing the pose can thus be informative for assessing the problems with traditional perception algorithms in this environment. COLMAP was capable of producing relatively low average reconstruction error across the scene as shown in Table~\ref{tab:pose_error}, enabling its use for pose estimation, but a number of adjustments were required for the algorithm to converge. First, COLMAP was not able to converge when provided with images alone and required initial approximate camera locations and orientations. In addition, particularly for the views in which the cameras were facing in the direction of the light source (terrain views 2 and 4), a substantial amount of parameter tuning was required after providing initial poses for COLMAP to successfully converge and produce a scene reconstruction. This process mostly involved adjustments to the feature matching and sparse reconstruction portions of the pipeline.\footnote{Appendix A in the \href{https://ti.arc.nasa.gov/dataset/PolarTrav/downloads/README.pdf}{dataset documentation} provides details on how the COLMAP reconstructions were produced.}

\begin{figure}[!htbp]
    \centering
    \includegraphics[width=0.45\textwidth]{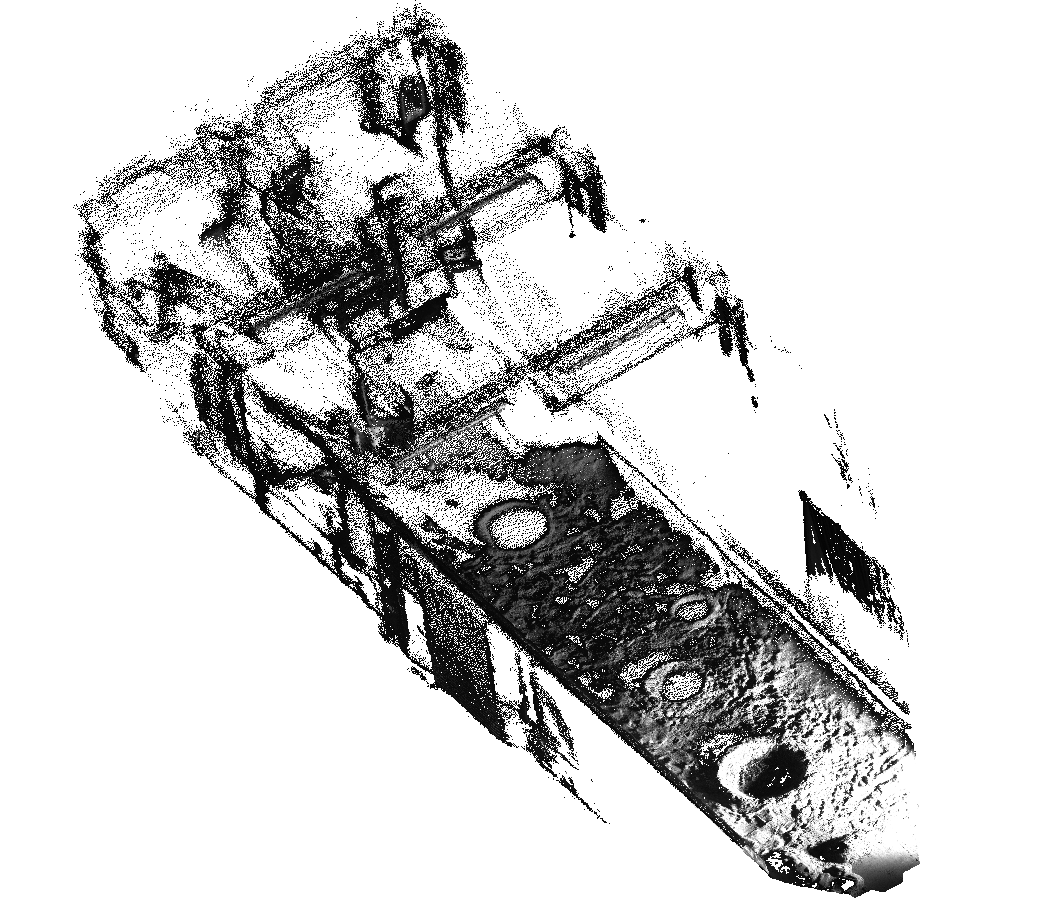}
    \caption{Dense reconstruction from COLMAP generated using images from terrain view 1. The scene reconstruction contains a number of points from the scene outside the test bed.}
    \label{fig:colmap_dense}
\end{figure}

Finally, the environment was not limited to the test bed for these scene reconstructions as the intent was to provide accurate pose. There are indeed many points external to the test bed - particularly for the portions of the scene farther along the test bed - that appear in the final dense reconstructions, as can be seen in the example in Figure~\ref{fig:colmap_dense}. These points may be more useful for feature matching as they come from human-made structures such as overhead lights that have sharp lines and corners not seen in natural terrain.

While COLMAP is thus capable of using this data to produce a reconstruction of the scene, it was only successful after significant manual adjustment of the parameters, in addition to initial pose estimation. The latter may be available for a rover traversing the lunar surface, but high sensitivity to parameter values indicates that this algorithm is not robust for the lunar polar environment.

\subsection{Dataset Limitations}

While effort was made to simulate a real lunar traverse, the images collected for this dataset are still limited by the conditions that exist in the lab, making them not quite representative of what can be expected on the surface of the moon. Limitations of the dataset include light fall-off, lack of motion blur, and presence of man-made materials and objects external to the test bed. Unlike on the lunar surface, where incident sunlight falls in rays that are approximately parallel, the proximity of the light source to the test bed means the amount of light reaching the terrain decreases according to the inverse square law as the cameras move farther away from the source. As a result, the appropriate exposure level changes as the cameras move across the scene, unlike what is expected for traversing actual lunar terrain.

Additionally, the cameras were stationary when images were recorded, so no motion blur is present as would be the case if recording images while the sensors were in motion. For certain applications – such as operating a rover under continuous motion – motion blur is expected to be present and would be more pronounced with longer exposure times. Finally, no attempt was made to post-process the images to mask out objects external to the test bed. Careful masking of the images  should be able to limit the effects of these visible objects when using the data.

\begin{table}[!tbp]
    \centering
    \begin{tabular}{c|c}
         View & Error (Pixels) \\
         \hline
         1 & 0.170 \\
         2 & 0.201 \\
         3 & 0.187 \\
         4 & 0.186 \\
         \hline
         Avg. & 0.186 \\
         \hline
    \end{tabular}
    \caption{Mean reconstruction error in pixels by terrain view from COLMAP.}
    \label{tab:pose_error}
\end{table}

\section{Conclusion}

The POLAR Traverse Dataset is designed to provide stereo pairs of images from a simulated traverse over lunar-like terrain under lunar polar lighting conditions for use developing robust perception algorithms and for better understanding of the lighting conditions of this environment. The dataset contains multiple traverses, parameterized by varying camera height and pitch, through four terrain views specified by the direction along the test bed in which the cameras were facing and the relative azimuth angle of incoming light. A total of 3,960 stereo pairs of images were collected across 24 total traverses, with images taken with 15 exposure times at each sample location. Ground truth geometry and camera position information are also available, the latter of which was generated using a multi-view stereo technique that can provide some insight into the issues these algorithms face in such an environment. These images should assist with algorithmic development and understanding of lighting conditions for upcoming missions to the lunar south polar regions.


\textbf{Acknowledgements} This work was supported by NASA Space Technology Graduate Research Opportunity (NSTGRO) grant 80NSSC22K1206. The Solar System Exploration Research Virtual Institute (SSERVI) has graciously provided access to their laboratory facilities for collecting this data. We would additionally like to thank Dr. Massimo Vespignani, Roshan Kalghatgi, and Molly O'Connor for their assistance with hardware setup, camera calibration, and terrain formation, respectively.

{
\bibliographystyle{IEEEtran}
\bibliography{polar-trav}

\begin{thebibliography}{10}
\providecommand{\url}[1]{#1}
\csname url@samestyle\endcsname
\providecommand{\newblock}{\relax}
\providecommand{\bibinfo}[2]{#2}
\providecommand{\BIBentrySTDinterwordspacing}{\spaceskip=0pt\relax}
\providecommand{\BIBentryALTinterwordstretchfactor}{4}
\providecommand{\BIBentryALTinterwordspacing}{\spaceskip=\fontdimen2\font plus
\BIBentryALTinterwordstretchfactor\fontdimen3\font minus \fontdimen4\font\relax}
\providecommand{\BIBforeignlanguage}[2]{{%
\expandafter\ifx\csname l@#1\endcsname\relax
\typeout{** WARNING: IEEEtran.bst: No hyphenation pattern has been}%
\typeout{** loaded for the language `#1'. Using the pattern for}%
\typeout{** the default language instead.}%
\else
\language=\csname l@#1\endcsname
\fi
#2}}
\providecommand{\BIBdecl}{\relax}
\BIBdecl

\bibitem{ennico-smith_viper_2022}
K.~Ennico-Smith, A.~Colaprete, D.~Lim, and D.~Andrews, ``The {VIPER} {Mission}, a {Resource}-{Mapping} {Mission} on {Another} {Celestial} {Body},'' in \emph{Space Resources Roundtable {XXII} Meeting}, Golden, CO, USA, Jun. 2022.

\bibitem{wong_characterization_2016}
U.~Wong, A.~Nefian, L.~Edwards, M.~Furlong, X.~Bouyssounouse, V.~To, M.~Deans, H.~Cannon, and T.~Fong, ``Characterization of {Stereo} {Vision} {Performance} for {Roving} at the {Lunar} {Poles},'' in \emph{{NASA} Exploration Science Forum 2016}, Moffett Field, CA, Jul. 2016.

\bibitem{moon2mars}
{NASA}, ``{NASA's Moon to Mars Strategy and Objectives Development},'' \url{https://go.nasa.gov/3zzSNhp}.

\bibitem{viper_rover}
NASA, ``{VIPER Mission Overview},'' \url{https://www.nasa.gov/viper/overview}.

\bibitem{shirley_viper_2022}
M.~{Shirley}, E.~{Balaban}, A.~{Colaprete}, R.~C. {Elphic}, H.~{Sanchez}, L.~{Falcone}, R.~{Beyer}, S.~{Banerjee}, and K.~{Bradner}, ``{VIPER Traverse Planning},'' in \emph{53rd Lunar and Planetary Science Conference}, ser. LPI Contributions, vol. 2678, Mar. 2022, p. 2874.

\bibitem{ArtemisScienceReport2021}
\BIBentryALTinterwordspacing
NASA, ``{Artemis III Science Definition Team Report},'' NASA, Tech. Rep. NASA/SP-20205009602, 2021. [Online]. Available: \url{https://www.nasa.gov/wp-content/uploads/2015/01/artemis-iii-science-definition-report-12042020c.pdf}
\BIBentrySTDinterwordspacing

\bibitem{fisher_evidence_2017}
\BIBentryALTinterwordspacing
E.~A. Fisher, P.~G. Lucey, M.~Lemelin, B.~T. Greenhagen, M.~A. Siegler, E.~Mazarico, O.~Aharonson, J.-P. Williams, P.~O. Hayne, G.~A. Neumann, D.~A. Paige, D.~E. Smith, and M.~T. Zuber, ``{Evidence for surface water ice in the lunar polar regions using reflectance measurements from the Lunar Orbiter Laser Altimeter and temperature measurements from the Diviner Lunar Radiometer Experiment},'' \emph{Icarus}, vol. 292, pp. 74--85, 2017. [Online]. Available: \url{https://www.sciencedirect.com/science/article/pii/S0019103516307795}
\BIBentrySTDinterwordspacing

\bibitem{hayne_micro_2020}
\BIBentryALTinterwordspacing
P.~O. Hayne, O.~Aharonson, and N.~Sch\"{o}rghofer, ``{Micro cold traps on the Moon},'' \emph{Nature Astronomy}, vol.~5, no.~2, pp. 169--175, Oct. 2020. [Online]. Available: \url{https://doi.org/10.1038/s41550-020-1198-9}
\BIBentrySTDinterwordspacing

\bibitem{crues_approaches_2023}
\BIBentryALTinterwordspacing
E.~Z. Crues, P.~Bielski, E.~Paddock, C.~Foreman, B.~Bell, C.~Raymond, T.~Hunt, and D.~Bulikhov, ``\BIBforeignlanguage{en}{Approaches for {Validation} of {Lighting} {Environments} in {Realtime} {Lunar} {South} {Pole} {Simulations}},'' in \emph{\BIBforeignlanguage{en}{2023 {IEEE} {Aerospace} {Conference}}}.\hskip 1em plus 0.5em minus 0.4em\relax Big Sky, MT, USA: IEEE, Mar. 2023, pp. 1--18. [Online]. Available: \url{https://ieeexplore.ieee.org/document/10115836/}
\BIBentrySTDinterwordspacing

\bibitem{null_identification_2023}
C.~H. Null, M.~K. Kaiser, T.~E. Wolters, J.~J. Marquez, A.~M. Cooter, and H.~C. Dischinger, ``\BIBforeignlanguage{en}{Identification of {Risks} to {EVA} {Created} by {Ambient} {Lighting} {Conditions} at the {Lunar} {South} {Pole}},'' in \emph{\BIBforeignlanguage{en}{12th {International} {Association} for the {Advancement} of {Space} {Safety} {Conference}}}, Osaka, Japan, May 2023.

\bibitem{stefano_fast_2004}
\BIBentryALTinterwordspacing
L.~D. Stefano, M.~Marchionni, and S.~Mattoccia, ``A fast area-based stereo matching algorithm,'' \emph{Image and Vision Computing}, vol.~22, no.~12, pp. 983--1005, Oct. 2004. [Online]. Available: \url{https://www.sciencedirect.com/science/article/pii/S0262885604000733}
\BIBentrySTDinterwordspacing

\bibitem{lunar_lab}
{SSERVI}, ``{Lunar Lab and Regolith Testbeds at NASA Ames},'' \url{https://sservi.nasa.gov/testbed/}.

\bibitem{isachenkov_char_2022}
\BIBentryALTinterwordspacing
M.~Isachenkov, S.~Chugunov, Z.~Landsman, I.~Akhatov, A.~Metke, A.~Tikhonov, and I.~Shishkovsky, ``Characterization of novel lunar highland and mare simulants for {ISRU} research applications,'' \emph{Icarus}, vol. 376, p. 114873, Apr. 2022. [Online]. Available: \url{https://www.sciencedirect.com/science/article/pii/S0019103521005108}
\BIBentrySTDinterwordspacing

\bibitem{long-fox_applicability_2022}
\BIBentryALTinterwordspacing
J.~Long-Fox, M.~P. Lucas, Z.~Landsman, C.~Millwater, D.~Britt, and C.~Neal, ``\BIBforeignlanguage{en}{Applicability of {Simulants} in {Developing} {Lunar} {Systems} and {Infrastructure}: {Geotechnical} {Measurements} of {Lunar} {Highlands} {Simulant} {LHS}-1},'' in \emph{\BIBforeignlanguage{en}{Proceedings of the 18th {Biennial} {International} {Conference} on {Engineering}, {Science}, {Construction}, and {Operations} in {Challenging} {Environments}}}, Denver, CO, USA, Apr. 2022, iSBN: 9780784484470. [Online]. Available: \url{https://ascelibrary.org/doi/epdf/10.1061/9780784484470.007}
\BIBentrySTDinterwordspacing

\bibitem{manta}
{Allied Vision}, ``{Manta G-419},'' \href{https://www.alliedvision.com/en/camera-selector/detail/manta/g-419/}{https://www.alliedvision.com/en/camera-selector/detail/manta/g-419/}.

\bibitem{tamron}
Tamron, ``{M111FM08},'' \href{https://www.tamron.vision/lenses/m111fm08/}{https://www.tamron.vision/lenses/m111fm08/}.

\bibitem{matlab}
{MATLAB}, ``{Using the Stereo Camera Calibrator App},'' \href{https://www.mathworks.com/help/vision/ug/using-the-stereo-camera-calibrator-app.html}{https://www.mathworks.com/help/vision/ug/using-the-stereo-camera-calibrator-app.html}.

\bibitem{schonberger_structure--motion_2016}
\BIBentryALTinterwordspacing
J.~L. Schonberger and J.-M. Frahm, ``\BIBforeignlanguage{en}{Structure-from-{Motion} {Revisited}},'' in \emph{\BIBforeignlanguage{en}{2016 {IEEE} {Conference} on {Computer} {Vision} and {Pattern} {Recognition} ({CVPR})}}.\hskip 1em plus 0.5em minus 0.4em\relax Las Vegas, NV, USA: IEEE, Jun. 2016, pp. 4104--4113. [Online]. Available: \url{http://ieeexplore.ieee.org/document/7780814/}
\BIBentrySTDinterwordspacing

\end{thebibliography}
}

\end{document}